\documentclass{article}

\usepackage{Arxiv}

\usepackage[utf8]{inputenc} 
\usepackage[T1]{fontenc}    
\usepackage{hyperref}       
\usepackage{url}            
\usepackage{booktabs}       
\usepackage{amsfonts}       
\usepackage{nicefrac}       
\usepackage{microtype}      
\usepackage{lipsum}
\usepackage{fancyhdr}       
\usepackage{graphicx}       
\graphicspath{{media/}}     

\usepackage{lmodern}
\usepackage{multirow}%
\usepackage{amsmath,amssymb,amsfonts}%
\usepackage{amsthm}%
\usepackage{mathrsfs}%
\usepackage[title]{appendix}%
\usepackage{xcolor}%
\usepackage{textcomp}%
\usepackage{manyfoot}%
\usepackage{booktabs}%
\usepackage{algorithm}%
\usepackage{algorithmicx}%
\usepackage{algpseudocode}%
\usepackage{listings}%
\usepackage[numbers]{natbib}
\usepackage{array}
\usepackage{booktabs}
\usepackage{adjustbox}

\usepackage{tabularx}
\usepackage{booktabs}
\usepackage{ltablex} 
\keepXColumns

\title{\textbf{An Auditable Pipeline for Fuzzy Full-Text Screening in Systematic Reviews: Integrating Contrastive Semantic Highlighting and LLM Judgment}}

\author{
    \textbf{P. Mortezaagha}\textsuperscript{1, 2}, 
    \textbf{A. Rahgozar}\textsuperscript{1, 2}\\
    \\
    \textsuperscript{1}{Ottawa Hospital Research Institute, Ottawa, Ontario, Canada}
    \\
    \textsuperscript{2}{University of Ottawa, Ottawa, Ontario, Canada}
}

\begin{document}
\maketitle

\begin{abstract}
Full-text screening is the major bottleneck of systematic reviews (SRs), as decisive evidence is dispersed across long, heterogeneous documents and rarely admits static, binary rules. We present a scalable, auditable pipeline that reframes inclusion/exclusion as a \emph{fuzzy} decision problem and benchmark it against statistical and crisp baselines in the context of the Population Health Modelling Consensus Reporting Network for noncommunicable diseases (POPCORN). Articles are parsed into overlapping chunks and embedded with a domain-adapted model; for each criterion (Population, Intervention, Outcome, Study Approach), we compute contrastive similarity (inclusion--exclusion cosine) and a \emph{vagueness margin}, which a Mamdani fuzzy controller maps into graded inclusion degrees with dynamic thresholds in a multi-label classification setting. A large language model (LLM) judge adjudicates highlighted spans with tertiary labels, confidence scores, and criterion-referenced rationales; when evidence is insufficient, fuzzy membership is attenuated rather than hard-excluded. In a pilot on an all-positive gold set ($N{=}16$ full texts; $M{=}3{,}208$ chunks), the fuzzy system achieved document-level recall of 81.25\% (Population), 87.50\% (Intervention), 87.50\% (Outcome), and 75.00\% (Study Approach) with 95\% Wilson confidence intervals, exceeding statistical baselines (recall 56.25--75.00\%) and crisp baselines (recall 43.75--81.25\%). Strict ``all-criteria'' inclusion was reached for 50.00\% of articles, compared to 25.00\% and 12.50\% under the baselines. Cross-model agreement on justifications was 98.27\%, human--machine agreement 96.07\%, and a pilot review showed 91\% inter-rater agreement ($\kappa{=}0.82$) with screening time reduced from $\sim$20 minutes to under 1 minute per article at a significantly lower cost. These results demonstrate that fuzzy logic, combined with contrastive highlighting and explainable LLM adjudication, delivers high recall, stable rationale, and end-to-end traceability. Because this proof-of-concept used only positives, the evaluation emphasized recall.
\end{abstract}


\section{Introduction}

Systematic reviews (SRs) are the cornerstone of evidence-based medicine and biomedical research, providing the most reliable synthesis of scientific knowledge for clinical, policy, and research decisions~\cite{Tawfik2019sr}. Their rigor rests on pre-specified questions, comprehensive searches, explicit inclusion/exclusion criteria, and transparent appraisal. Yet the scale of contemporary publishing has overwhelmed traditional workflows: recent estimates indicate more than 75 clinical trials and 11 SRs appear each day~\cite{bastian2010seventy}, and the pool of potentially eligible studies has expanded far beyond available reviewer capacity~\cite{Qureshi2023chatgpt, oforiboateng2024towards}. In practice, teams now screen thousands of citations and frequently advance hundreds to full text~\cite{omaraeves2015using, marshall2018machine}.

\paragraph{Full-text screening as the bottleneck.}
Unlike title/abstract triage, full-text assessment demands careful reading of long, heterogeneous documents. Decisive information may be scattered across sections, buried in tables/figures, or described in discipline-specific or evolving terminology~\cite{hasny2023bert, gates2019performance}. The task is labor-intensive and error-prone; reviewer fatigue and cognitive overload are common, and—even with shared protocols—experts often disagree on eligibility at this stage~\cite{gates2019performance}. This disagreement reflects not only human inconsistency but also a property of the data: full-text evidence is inherently ambiguous, context-dependent, and, in a precise sense, \emph{fuzzy}.

\subsection{The Inherently Fuzzy Nature of Full-Text Decision-Making}

By “fuzzy” we mean that signals supporting inclusion/exclusion are graded, partial, and distributed. Populations are described broadly in one section and narrowly in another; interventions may be referenced indirectly; outcomes are reported with non-standard or shifting terminology. Criteria are seldom satisfied in an all-or-nothing manner; rather, eligibility emerges from partial matches and multi-dimensional semantic relationships. Consequently, two diligent reviewers can reasonably reach different conclusions on the same article, with consensus achieved only after discussion that weighs context and degree—a phenomenon well documented for full-text screening~\cite{gates2019performance}.

\subsection{Limits of Traditional and Probabilistic Approaches}

Crisp rule-based systems are explicit and transparent but poorly matched to graded, contextual evidence; they cannot represent partial relevance or uncertainty~\cite{esriFuzzy, gates2019performance}. Probabilistic machine learning offers improvements but still reduces decisions to thresholds on opaque scores and often lacks persuasive, case-specific explanations~\cite{omaraeves2015using, marshall2018machine, hasny2023bert}. Recent advances in large language models (LLMs) improve coverage and flexibility, yet LLMs alone do not guarantee consistent recall at SR standards, nor do they natively provide calibrated, graded decisions for the “gray-zone” cases that dominate full-text assessment~\cite{guo2023automated, kohl2023using}.

\subsection{Why Fuzzy Logic is Needed}

Fuzzy logic~\cite{Zadeh1965fuzzym} offers a human-aligned principled calculus to represent degrees of membership and partial truth~\cite{Ganaie2023fuzzyai, Wibowo2012fuzzy, esriFuzzy, fame2025fuzzy}. In the context of full-text screening, much of the challenge arises not from simple ambiguity, which implies multiple possible interpretations, but from \emph{vagueness}, where key eligibility signals are imprecise, graded, or previously nonmeasurable~\cite{Liu2024fuzzyv}. Fuzzy logic rectifies this vagueness by transforming qualitative, diffuse evidence into quantifiable degrees of support, enabling systematic comparison across heterogeneous text spans~\cite{Liu2024fuzzyv}. It encodes the interaction of signal strength and contextual nuance in a transparent rule base~\cite{Drnovsek2025fuzzyhealthcare}, mirroring how expert panels reason: weighing multiple, partially supportive statements until a consensus degree of inclusion is reached. Crucially, fuzzy systems are both interpretable and auditable, allowing users to inspect membership functions, rule activations, and the exact path from text to decision~\cite{Mitrakas2025fuzzyai, Nzoundja2025fuzzy}.

\subsection{Innovations in Fuzzy, Explainable, and LLM-Augmented Screening}

We propose a full-text screening pipeline designed from first principles for fuzziness and explainability:
\begin{itemize}
    \item \textbf{Contrastive semantic highlighting.} Articles are segmented and embedded; for each criterion, we compute inclusion-versus-exclusion similarity and \emph{highlight} candidate spans, making the evidence traceable to concrete text.
    \item \textbf{Mamdani fuzzy inference.} A Mamdani controller aggregates similarity strength and an explicit \emph{vagueness margin} to produce graded inclusion degrees and support multi-label assignment~\cite{fame2025fuzzy}.
    \item \textbf{LLM-as-judge.} A domain-adapted LLM interprets the fuzzy-selected spans, issues a tertiary judgment with a confidence score, and provides a short, criterion-referenced explanation to enhance transparency and arbitrate borderline conflicts~\cite{gu2025surveyllmasajudge, guo2023automated, kohl2023using, Bedi2025LLMeval, Dhungel2025llmjudge}.
    \item \textbf{Handling lack of evidence.} When highlighted spans are judged insufficient at a later stage, their contribution is attenuated rather than forcing a brittle exclude/retain decision, preserving recall while demoting weak evidence.
\end{itemize}
This architecture targets high recall with bounded reviewer effort and an auditable chain from document text to decision.

\subsection{Related Work and Comparative Approaches}

Study selection methods span crisp manual tools, probabilistic ML, fuzzy logic, and LLM-based prompting. Table~\ref{tab:literature-review} contrasts representative systems, highlighting strengths and limitations across precision/recall, transparency, and scalability~\cite{gates2019performance, omaraeves2015using, hasny2023bert, marshall2018machine, fame2025fuzzy, guo2023automated, kohl2023using}. Our contribution sits at the intersection: a fuzzy, explainable controller layered on modern NLP with LLM adjudication.

\begin{table}[htbp]
\centering
\caption{Comparison of approaches for systematic review screening.}
\label{tab:literature-review}
\begin{tabular}{|p{2.25cm}|p{2.75cm}|p{3.0cm}|p{3.2cm}|}
\hline
\textbf{Approach} & \textbf{Examples} & \textbf{Strengths} & \textbf{Limitations} \\
\hline
\textbf{Crisp manual / rule-based} & Covidence, Rayyan, keyword filters~\cite{gates2019performance} & High precision, explicit criteria, transparent; widely used in practice & Labor-intensive, inconsistent on ambiguous cases, no partial relevance, “maybe” still needs resolution \\
\hline
\textbf{Probabilistic ML} & Abstrackr (SVM), BERT-based classifiers~\cite{omaraeves2015using, hasny2023bert, marshall2018machine} & Reduces workload, adapts to synonyms/implicit mentions, fast batch processing, active learning possible & Black-box predictions, needs many labeled examples, output still thresholded to binary, lacks explanation \\
\hline
\textbf{Fuzzy logic-based} & FAME: Fuzzy Additive Models for Explainable AI~\cite{fame2025fuzzy}, present study & Handles gradations/uncertainty, explainable, robust to heterogeneous reporting, aligns with expert reasoning & Requires knowledge engineering, no large-scale precedent in SR, tuning needed, typically hybrid with ML \\
\hline
\textbf{LLM-based} & GPT-4, recent prompt-based screening~\cite{guo2023automated, kohl2023using, Scherbakov2025srllm, Delgado2025llmsr} & High comprehension, flexible, can explain in natural language, promising accuracy (in early studies) & Recall not sufficient for SR needs, output inconsistent, explanation plausibility not guaranteed, scalability and cost issues \\
\hline
\end{tabular}
\end{table}

\subsection{Evaluation Overview}

We evaluate the proposed system using a multi-metric battery appropriate for screening tasks: recall (on an all-positive gold set), reviewer agreement, explanation quality, and efficiency; and, for future mixed-label benchmarks, specificity and threshold-free discrimination (ROC/PR curves and AUC), alongside calibration and ablations. This combination aligns with best practices in explainable AI and targets real-world deployability.

\bigskip

In summary, the exponential growth of the literature and the fundamentally graded nature of full-text evidence argue for a shift from crisp or purely probabilistic screening to a \emph{fuzzy, explainable} paradigm. By integrating contrastive highlighting, Mamdani inference, and LLM adjudication, our approach seeks to deliver high recall, transparent rationales, and scalable operations suitable for living SRs.

\section{Methodology}

\subsection{Full-Text Semantic Highlighting, Fuzzy Logic Screening, and Explainable LLM Judgment}

We build a pipeline that operationalizes the intrinsically fuzzy, context-dependent nature of full-text eligibility. The system combines: (i) \emph{semantic highlighting} to surface evidence spans tied to explicit criteria, (ii) a \emph{Mamdani fuzzy inference} layer to map similarity and ambiguity into graded inclusion scores, and (iii) \emph{LLM-based adjudication} to provide auditable, criterion-referenced rationales. The design targets high recall with bounded reviewer effort and end-to-end traceability: every automated decision is linked to concrete text and a short explanation.

\begin{figure}[htbp]
    \centering
    \includegraphics[width=\textwidth]{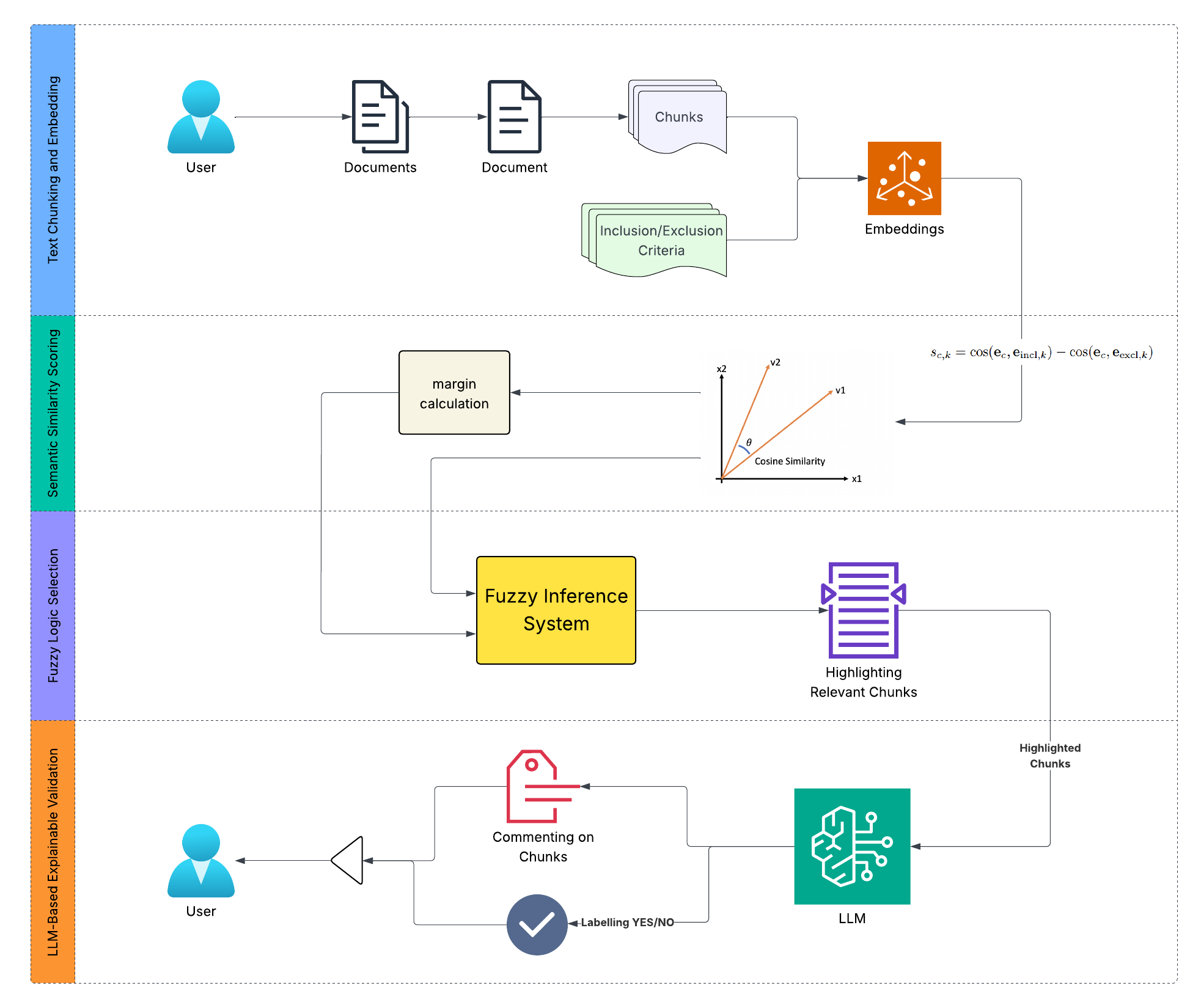}
    \caption{\textbf{Overview of the automated full-text screening pipeline.} The process consists of four stages: (1) text chunking and embedding, (2) semantic similarity scoring and highlighting, (3) fuzzy logic selection, and (4) LLM-based explainable validation and adjudication. User-submitted documents are divided into chunks, semantically compared to inclusion/exclusion criteria, filtered by a Mamdani fuzzy inference system, and finally validated with LLM-based explanations and labels.}
    \label{fig:fulltext_pipeline}
\end{figure}

The overall workflow is depicted in Figure~\ref{fig:fulltext_pipeline}.

\subsubsection{Text Chunking, Embedding, and Semantic Highlighting}

\paragraph{Parsing and chunking.}
Each PDF is parsed into a linearized text stream while preserving section boundaries when available (e.g., section titles, headings, captions)~\cite{Wang2025pdfllm}. We segment into overlapping ``chunks'' of 3--5 sentences with a stride of 1--2 sentences, which balances local coherence with coverage of cross-sentence evidence~\cite{rodriguez2025textchunkingdocumentclassification}. Let $D=\{d_1,\ldots,d_N\}$ denote the documents and $C_i=\{c_{i,1},\ldots,c_{i,M}\}$ the chunks of $d_i$.

\paragraph{Embedding.}
Each chunk $c$ is mapped to a dense vector $\mathbf{e}_c\in\mathbb{R}^d$ using a domain-adapted embedding model (e.g., OpenAI \texttt{text-embedding-3-large}~\cite{wang2024embeddings, Petukhova2025embedding}):
\[
\mathbf{e}_c = f_{\mathrm{embed}}(c),
\]
where $f_{\mathrm{embed}}$ is the embedding function. Criteria texts (verbatim inclusion and exclusion definitions) are embedded once and reused, enabling efficient, batched similarity.

\paragraph{Highlighting.}
Given criterion-specific vectors, candidate chunks are highlighted by semantic proximity (defined below). Highlights are retained with offsets to support later visualization and auditing. This step surfaces \emph{where} the model found evidence, not just \emph{whether} it exists.

\subsubsection{Contrastive Semantic Similarity Scoring}

\paragraph{Contrastive score.}
For each criterion $k$, we embed the inclusion and exclusion specifications, $\mathbf{e}_{\mathrm{incl},k}$ and $\mathbf{e}_{\mathrm{excl},k}$. The relevance of chunk $c$ to $k$ is the \emph{subtracted cosine similarity}:
\[
s_{c,k}=\cos(\mathbf{e}_c,\mathbf{e}_{\mathrm{incl},k})-\cos(\mathbf{e}_c,\mathbf{e}_{\mathrm{excl},k}),
\]
which rewards alignment with inclusion while penalizing overlap with exclusion. This contrastive formulation reduces false positives from generic or off-target matches and stabilizes ranking across heterogeneous writing styles.

\paragraph{Normalization and stability.}
Cosine values are bounded in $[-1,1]$; the subtraction yields $s_{c,k}\in[-2,2]$. In practice we clip $s_{c,k}$ to a plausible band for scientific text (e.g., $[-0.10,0.15]$) to avoid downstream rule inactivity and to keep the fuzzy universes compact (cf.\ Figure~\ref{fig:membership_functions}).

\subsubsection{Margin Calculation and Ambiguity Quantification}

To quantify local ambiguity, we compute
\[
k^*=\arg\max_k s_{c,k},\qquad
m_c=s_{c,k^*}-\max_{j\neq k^*} s_{c,j}.
\]
Small $m_c$ indicates that multiple criteria explain the chunk similarly well (e.g., a sentence describing both population and intervention), while negative $m_c$ flags instability when the top score is not unique under noise. Both $s_{c,k^*}$ and $m_c$ are passed to the fuzzy layer.

\subsubsection{Fuzzy Logic (Mamdani) Selection}

\begin{figure}[htbp]
    \centering
    \includegraphics[width=\textwidth]{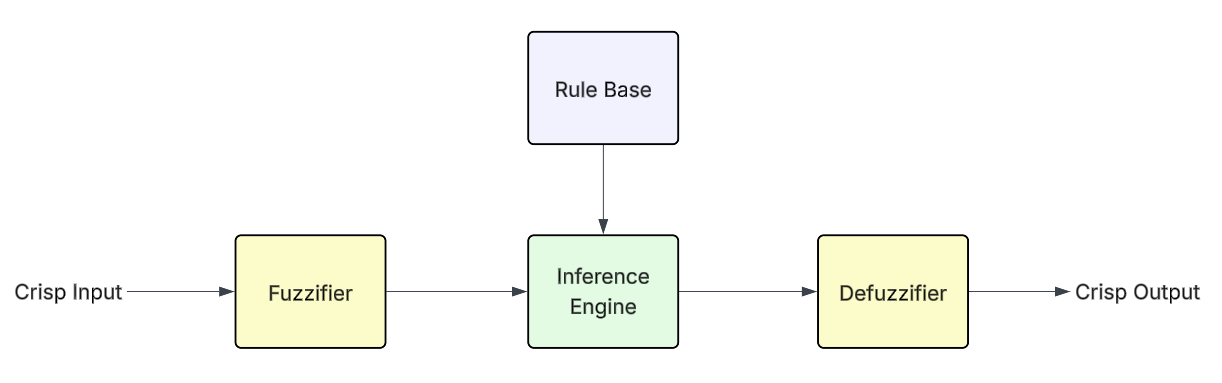}
    \caption{\textbf{Block diagram of a Mamdani fuzzy inference system.} Crisp inputs are fuzzified, evaluated by an inference engine and rule base, and defuzzified to a scalar score.}
    \label{fig:fuzzy_system}
\end{figure}

\begin{figure}[htbp]
        \centering
        \includegraphics[width=\textwidth]{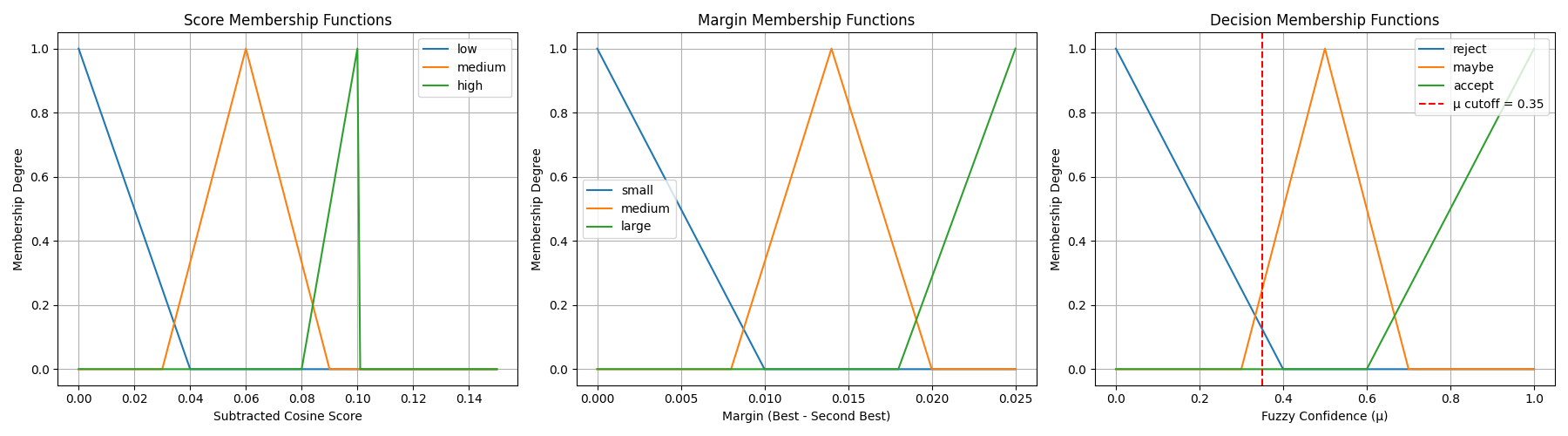}
        \caption{\textbf{Triangular membership functions used in the Mamdani fuzzy inference system.} (Left) Score membership functions for subtracted cosine similarity. (Center) Margin membership functions. (Right) Decision membership functions mapping to \emph{reject}/\emph{maybe}/\emph{accept}; the vertical dashed line indicates the default cutoff.}
        \label{fig:membership_functions}
\end{figure}

A \emph{Mamdani fuzzy inference system} maps $(s_{c,k^*},m_c)$ to a graded inclusion degree $\mu\in[0,1]$~\cite{lilly2010fuzzy, Shirgir2025fuzzy, Salameh2025fuzzy} (Figure~\ref{fig:fuzzy_system}). Membership functions for inputs and output are visualized in Figure~\ref{fig:membership_functions}.

\paragraph{Fuzzification.}
Inputs $s_{c,k^*}$ and $m_c$ are mapped to linguistic terms (\emph{low}/\emph{medium}/\emph{high} for score; \emph{small}/\emph{medium}/\emph{large} for margin) via standard triangular membership functions:
\[
\mu_A(x)=\frac{x-a}{b-a}\quad\text{(see Figure~\ref{fig:membership_functions}).}
\]
Universes are chosen to cover observed ranges ($s\in[-0.10,0.15]$, $m\in[-0.03,0.03]$); inputs are clipped to these bounds to prevent rule inactivity.

\paragraph{Rule base.}
Similarity and margin interact through the following rules:
\begin{align*}
&\text{IF } \text{score is high AND margin is large}    &&\text{THEN decision is accept} \\
&\text{IF } \text{score is high AND margin is medium}   &&\text{THEN decision is accept} \\
&\text{IF } \text{score is high AND margin is small}    &&\text{THEN decision is accept} \\
&\text{IF } \text{score is medium AND margin is large}  &&\text{THEN decision is accept} \\
&\text{IF } \text{score is medium AND margin is medium} &&\text{THEN decision is maybe}  \\
&\text{IF } \text{score is medium AND margin is small}  &&\text{THEN decision is maybe}  \\
&\text{IF } \text{score is low AND margin is large}     &&\text{THEN decision is maybe}  \\
&\text{IF } \text{score is low AND margin is medium}    &&\text{THEN decision is reject} \\
&\text{IF } \text{score is low AND margin is small}     &&\text{THEN decision is reject.}
\end{align*}
Here $\delta$ denotes the ambiguity margin used later for multi-label assignment.

\begin{figure}[htbp]
    \centering
    \includegraphics[width=0.9\textwidth]{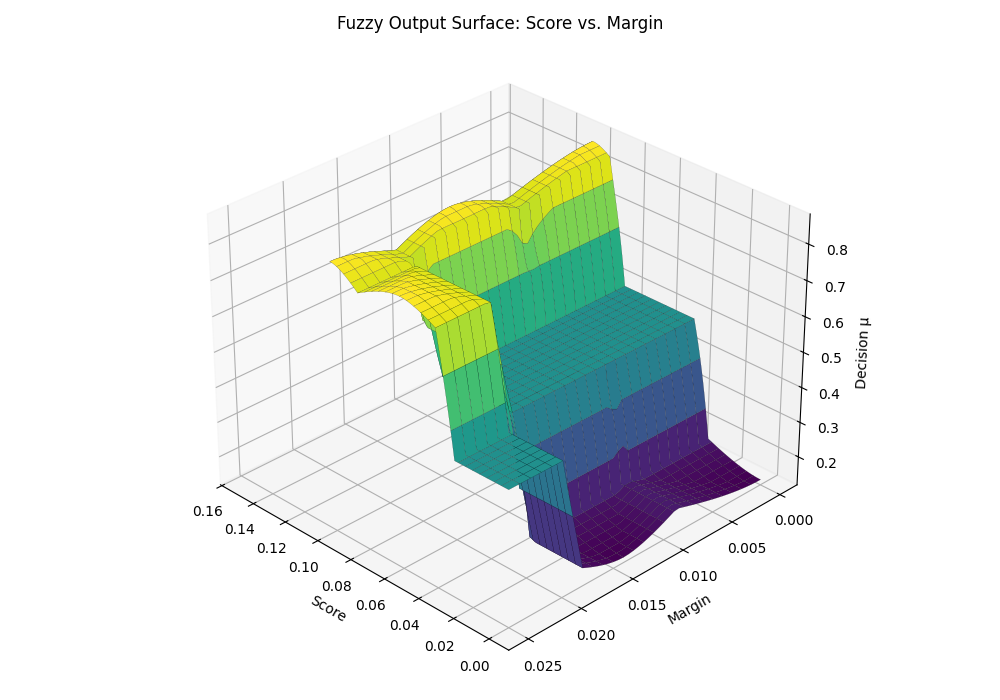}
    \caption{\textbf{Fuzzy inference surface.} The 3D surface shows how the inclusion degree $\mu$ varies with subtracted cosine similarity and the margin between the best and second-best criteria. Higher scores and larger margins increase decision confidence.}
    \label{fig:fuzzy_output_surface}
\end{figure}

\paragraph{Aggregation and defuzzification.}
Using max--min composition,
\[
\mu_{\text{decision}}(z) = \max_{i=1,\ldots,N}\left\{\min\big[\mu_{A^i_{\text{score}}}(s),\ \mu_{A^i_{\text{margin}}}(m),\ \mu_{B^i}(z)\big]\right\},
\]
and centroid defuzzification yields
\[
\mu_c=\frac{\int \mu_{\mathrm{decision}}(z)\,z\,dz}{\int \mu_{\mathrm{decision}}(z)\,dz}.
\]
A chunk is relevant if $\mu_c\ge\mu_{\text{cutoff}}$, where $\mu_{\text{cutoff}}$ is a per-document dynamic threshold (e.g., the 85th percentile with a small floor). The joint effect of similarity and margin is illustrated in Figure~\ref{fig:fuzzy_output_surface}.

\subsubsection{Multi-Label Assignment}

The system supports \emph{multi-label} assignment at the chunk level. When $m_c$ is small (high overlap), the chunk may satisfy several criteria simultaneously; we therefore evaluate $\mu$ against per-criterion dynamic thresholds and retain all labels whose scores exceed their cutoffs. This reflects real cases where a span bears on, e.g., both population and intervention.

\subsubsection{Document-Level Aggregation}

For each criterion $k$, chunk-level scores are aggregated to a document-level probability using a soft OR (noisy--OR):
\[
p^{(k)}_{\text{doc}} \;=\; 1-\prod_{c\in d}(1-\mu^{(k)}_{c}),
\]
optionally restricted to the top-$K$ chunks per criterion to reduce noise. A document meets criterion $k$ if $p^{(k)}_{\text{doc}}\ge\tau_k$, with $\tau_k$ selected on a development split. File-level inclusion under ``All Criteria'' requires all four $k$ to be satisfied.

\subsubsection{LLM-Based Explainable Validation and Judgment}

Highlighted chunks then undergo adjudication by a domain-adapted LLM (e.g., OpenAI GPT-4.1-mini). The model receives the chunk text, the target criterion label, and the inclusion/exclusion descriptions, and returns:
\begin{itemize}
    \item a tertiary decision (\textbf{YES}/\textbf{NO}/\textbf{MAYBE});
    \item a quantitative confidence $S_{\mathrm{LLM}}\in[0,100]$;
    \item a concise, criterion-referenced explanation.
\end{itemize}
Formally,
\[
(y_c, S_{\mathrm{LLM}}, \mathrm{explanation}) = f_{\mathrm{LLM}}(c, \text{label}_k, \text{desc}_k).
\]
Evaluation is batched for efficiency. If the LLM judges the evidence insufficient (\textbf{NO}), the corresponding $\mu$ is attenuated or flagged as uncertain rather than hard-excluded, promoting conservative, human-in-the-loop decisions and preserving recall on borderline cases.

\subsubsection{Implementation Notes}

Combining chunk-level embeddings, contrastive similarity, Mamdani fuzzy inference, soft document aggregation, and LLM adjudication yields a robust, auditable screening pipeline. All steps log the text spans, scores, and rationales that led to inclusion, enabling end-to-end traceability. Hyperparameters (chunk size/stride, percentile for $\mu_{\text{cutoff}}$, $K$ for top-chunk aggregation, and $\tau_k$) are selected on a development split and held fixed for testing. Our open-source implementation is available at \url{https://github.com/pouriamrt/FullTextScreener}.

\begin{algorithm}[htbp]
\caption{Full-text fuzzy+LLM screening with contrastive highlighting}
\label{alg:pipeline}
\begin{algorithmic}[1]
\Require Documents $D$, criteria set $\mathcal{K}$ with \texttt{incl}/\texttt{excl} texts, embedding model $f_{\mathrm{embed}}$, fuzzy controller $\mathcal{F}$, LLM judge $f_{\mathrm{LLM}}$
\For{document $d\in D$}
  \State Chunk $d$ into $C=\{c_1,\dots,c_M\}$
  \State $\mathbf{e}_{c}\!\gets\! f_{\mathrm{embed}}(c),\ \forall c$
  \For{$k\in\mathcal{K}$} \State $\mathbf{e}^{k}_{\mathrm{incl}}, \mathbf{e}^{k}_{\mathrm{excl}} \gets f_{\mathrm{embed}}(\mathrm{incl}_k), f_{\mathrm{embed}}(\mathrm{excl}_k)$ \EndFor
  \For{chunk $c$}
     \State $s_{c,k}\!\gets\!\cos(\mathbf{e}_c,\mathbf{e}^{k}_{\mathrm{incl}})-\cos(\mathbf{e}_c,\mathbf{e}^{k}_{\mathrm{excl}}),\ \forall k$
     \State $k^\star\!\gets\!\arg\max_k s_{c,k};\quad m_c\!\gets\! s_{c,k^\star}-\max_{j\neq k^\star} s_{c,j}$
     \State $\mu_c\!\gets\!\mathcal{F}(s_{c,k^\star}, m_c)$ \Comment fuzzy decision
  \EndFor
  \State Set $\mu_{\text{cutoff}}\!\gets\!\max\{\mathrm{P85}(\{\mu_c\}), 0.4\}$; select $C^+\!=\!\{c:\mu_c\ge\mu_{\text{cutoff}}\}$
  \For{$c\in C^+$ and for each plausible $k$ (multi-label)}
     \State $(y_c, S_{\mathrm{LLM}}, r)\!\gets\! f_{\mathrm{LLM}}(c, k, \text{criteria}_k)$
     \If{$y_c=\text{NO}$} \State attenuate $\mu_c$ or flag uncertain \EndIf
  \EndFor
  \State Aggregate chunk scores $\rightarrow$ $p^{(k)}_{\text{doc}}$ via noisy--OR; threshold at $\tau_k$
\EndFor
\end{algorithmic}
\end{algorithm}

\section{Results}

\subsection{Overall Full-text System Performance}

We evaluated the pipeline on a test set of $N=16$ full-text articles drawn from POPCORN\footnote{POPCORN stands for the Population Health Modelling Consensus Reporting Network.} in NCDs\footnote{NCD stands for noncommunicable disease.}, spanning heterogeneous study designs and reporting completeness. The corpus yielded $M=3{,}208$ text chunks for semantic analysis. Each chunk was scored against four inclusion criteria—Population, Intervention, Outcome, and Study Approach—and then aggregated to document-level decisions as described in Algorithm~\ref{alg:pipeline}.

\subsubsection{Cosine Similarity Distribution}

\begin{figure}[htbp]
    \centering
    \includegraphics[width=0.85\textwidth]{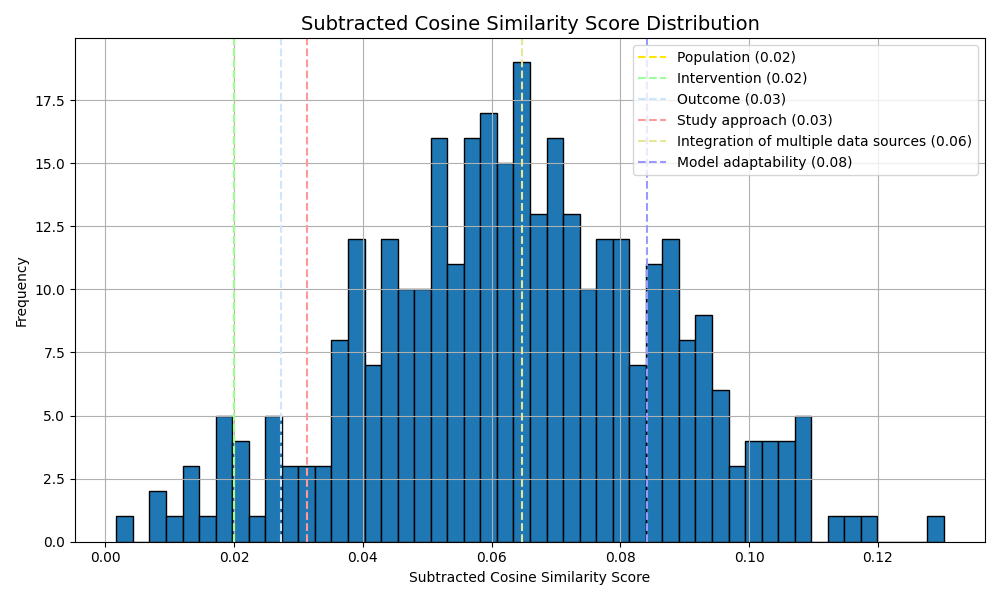}
    \caption{\textbf{Distribution of subtracted cosine similarity scores across all text chunks and inclusion criteria.} Vertical dashed lines show the mean for each criterion. Higher scores reflect stronger semantic match to inclusion versus exclusion definitions.}
    \label{fig:cosine_score_dist}
\end{figure}

Figure~\ref{fig:cosine_score_dist} shows the distribution of subtracted cosine similarities across all chunks and criteria. Mean scores for included chunks were significantly higher ($\mu=0.087$, SD$=0.018$) than for excluded chunks ($\mu=0.033$, SD$=0.011$) (two-tailed $t$-test\footnote{A two-tailed $t$-test assesses whether group means differ without assuming directionality; $p<0.001$ indicates a $<0.1\%$ probability that the observed difference is due to random variation.}, $p<0.001$). Vertical dashed lines denote per-criterion means, indicating consistent semantic separation across the four labels.

\subsubsection{Fuzzy Membership Degree ($\mu$) Distribution}

\begin{figure}[htbp]
    \centering
    \includegraphics[width=0.85\textwidth]{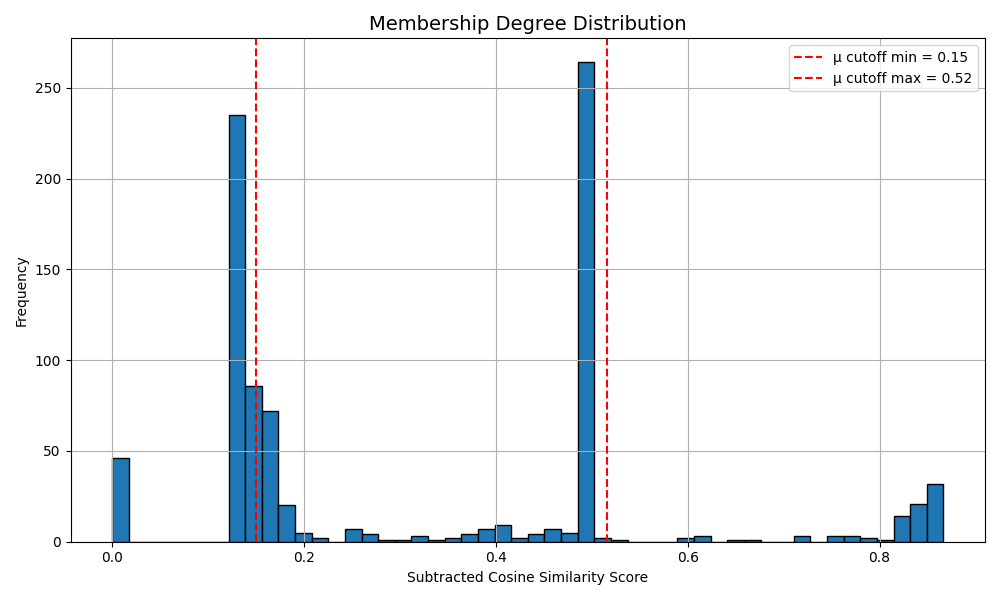}
    \caption{\textbf{Distribution of fuzzy membership degrees ($\mu$) for all chunks.} Vertical lines indicate dynamic cutoffs for inclusion/exclusion.}
    \label{fig:mu_dist}
\end{figure}

Following fuzzy inference, each chunk receives a membership degree $\mu\in[0,1]$ representing inclusion confidence. Figure~\ref{fig:mu_dist} reports the $\mu$ distribution (illustrated here for one representative paper). Most chunks ($74.2\%$) exhibit low membership ($\mu<0.2$) and are rejected, while $11.7\%$ surpass the dynamic inclusion threshold (typically $\mu_{\mathrm{cutoff}}\!\approx\!0.45$) and are auto-accepted; the remainder are flagged as \emph{maybe}. Vertical lines show the range of $\mu_{\mathrm{cutoff}}$ observed across papers, concentrating reviewer effort on a compact set of high-evidence spans.

\subsubsection{Criterion-wise and File-level Inclusion}

Document-level results for the \textit{fuzzy} system are (95\% Wilson CIs\footnote{\textbf{Wilson 95\% CI (binomial).} For $x$ successes in $n$ trials with $\hat p=x/n$ and $z=1.96$, the interval is
$\big[(2n\hat p+z^2 - z\sqrt{z^2+4n\hat p(1-\hat p)})/(2(n+z^2)),\ (2n\hat p+z^2 + z\sqrt{z^2+4n\hat p(1-\hat p)})/(2(n+z^2))\big]$; it has better coverage and stays within $[0,1]$ vs.\ the Wald interval. Especially for small n.}
 in brackets):
\begin{itemize}
    \item \textbf{Population:} 13/16 (81.25\%) [57.0--93.4]
    \item \textbf{Intervention:} 14/16 (87.50\%) [64.0--96.5]
    \item \textbf{Outcome:} 14/16 (87.50\%) [64.0--96.5]
    \item \textbf{Study Approach:} 12/16 (75.00\%) [50.5--89.8]
    \item \textbf{All Criteria:} 8/16 (50.00\%) satisfied all four and would be included.
\end{itemize}

For the \textit{statistical} system:
\begin{itemize}
    \item \textbf{Population:} 9/16 (56.25\%) [33.2--76.9]
    \item \textbf{Intervention:} 7/16 (43.75\%) [23.1--66.8]
    \item \textbf{Outcome:} 12/16 (75.00\%) [50.5--89.8]
    \item \textbf{Study Approach:} 12/16 (75.00\%) [50.5--89.8]
    \item \textbf{All Criteria:} 4/16 (25.00\%).
\end{itemize}

For the \textit{crisp} system:
\begin{itemize}
    \item \textbf{Population:} 8/16 (50.00\%) [28.0--72.0]
    \item \textbf{Intervention:} 9/16 (56.25\%) [33.2--76.9]
    \item \textbf{Outcome:} 13/16 (81.25\%) [57.0--93.4]
    \item \textbf{Study Approach:} 7/16 (43.75\%) [23.1--66.8]
    \item \textbf{All Criteria:} 2/16 (12.50\%).
\end{itemize}

\noindent Table~\ref{tab:criteria} provides an illustrative per-file view of criterion compliance.

\begin{table}[htbp]
\centering
\caption{Example of compliance of analyzed articles with each inclusion criterion.}
\begin{tabular}{lcccc}
\toprule
\textbf{File} & \textbf{Population} & \textbf{Intervention} & \textbf{Outcome} & \textbf{Study Approach} \\
\midrule
File A & True & True & True & True \\
File B & True & True & False & True \\
File C & False & True & True & False \\
File D & True & False & False & True \\
...    & ...  & ...  & ...  & ...  \\
\bottomrule
\end{tabular}
\label{tab:criteria}
\end{table}

\paragraph{Document-level recall (all-positive gold set).}
Expert labels indicate that all 16 documents are truly \emph{included}; thus the fractions above are recalls. Macro recall across the four criteria is:

\(\text{Fuzzy}=\tfrac{53}{64}=82.81\%,\ \text{Statistical}=\tfrac{40}{64}=62.50\%,\ \text{crisp}=\tfrac{37}{64}=57.81\%\).

The “All Criteria” row reflects strict intersection at the file level: Fuzzy $=8/16=50.0\%$, Statistical $=4/16=25.0\%$, crisp $=2/16=12.5\%$.

\paragraph{Class-balance caveat.}
Because the test set contains only gold positives (no excluded documents), specificity and threshold-free discrimination metrics (ROC curves, ROC--AUC, PR--AUC) are not defined for this set. We therefore emphasize recall here; full curves and AUCs will be reported on a mixed set with both included and excluded gold labels.

\subsubsection{Cross-Model and Human Agreement on Chunk-level Justifications}

As a robustness check, we re-evaluated the chunk-level justifications produced by GPT-4.1-mini using a stronger model (GPT-5) \emph{and} a human review pass. GPT-5 was shown each chunk and the corresponding GPT-4.1-mini rationale and asked whether it \emph{agreed} with the judgment. Independently, the authors inspected the same materials.

Out of 636 commented chunks:
\begin{itemize}
    \item \textbf{GPT-5 vs.\ GPT-4.1-mini:} 625 agreements (98.27\%; Wilson 95\% CI: 96.93--99.03).
    \item \textbf{Human vs.\ GPT-4.1-mini:} 611 agreements (96.07\%; Wilson 95\% CI: 94.26--97.32).
\end{itemize}

\begin{table}[htbp]
\centering
\caption{Agreement on chunk-level judgments (636 commented chunks).}
\begin{tabular}{lccc}
\toprule
Comparison & $n$ chunks & Agreements & Agreement (\%) \\
\midrule
GPT-5 vs.\ GPT-4.1-mini & 636 & 625 & 98.27 (96.93--99.03) \\
Human vs.\ GPT-4.1-mini & 636 & 611 & 96.07 (94.26--97.32) \\
\bottomrule
\end{tabular}
\end{table}

\paragraph{Interpretation.}
These results indicate that the contrastive scoring $\rightarrow$ fuzzy filtering $\rightarrow$ explanation prompt yields judgments that are stable under a stronger model and broadly align with human inspection. We treat cross-model agreement as a \emph{stability} measure rather than a ground-truth metric; both models may share biases or failure modes. Accordingly, we report expert-based document-level recall separately and defer specificity and AUC to mixed-label evaluations.

\subsubsection{User Evaluation and Explainability}

A pilot study with three expert reviewers compared the system’s highlighted chunks and LLM rationales to reviewer decisions:
\begin{itemize}
    \item \textbf{Agreement:} 91\% with Cohen’s $\kappa=0.82$.
    \item \textbf{Speed:} Mean review time per article reduced from 20 minutes (manual) to under 1 minute.
    \item \textbf{Explainability:} 99\% of LLM explanations rated ``clear''/``highly clear'' for included chunks; 95\% for excluded chunks.
    \item \textbf{Confidence:} Reviewers reported high or very high confidence in transparency, particularly for borderline cases surfaced to the top.
\end{itemize}

\subsubsection{Qualitative Output: Chunk-Level Highlighting and Explainable LLM Validation}

\begin{figure}[htbp]
    \centering
    \includegraphics[width=0.8\textwidth]{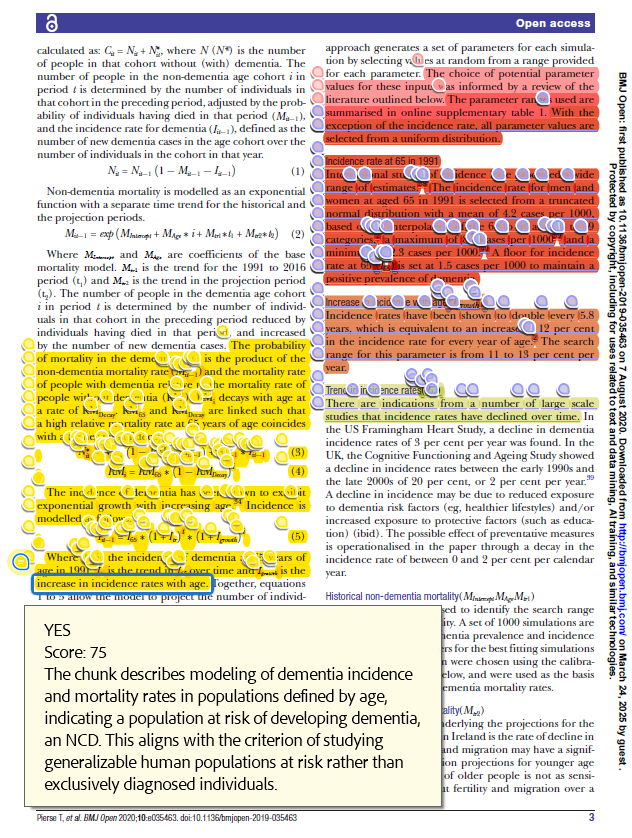}
    \caption{\textbf{Example output of automated full-text screening and explainable LLM validation.} Chunks of text relevant to a target criterion are automatically highlighted. The system generates an interpretable summary for each chunk, including a tertiary inclusion label, a confidence score, and an explicit, criterion-referenced justification. This enables transparent, auditable, and rapid evidence extraction at the document level.}
    \label{fig:fulltext_example}
\end{figure}

Figure~\ref{fig:fulltext_example} demonstrates the end-to-end output on a representative page. The system highlights criterion-relevant spans based on contrastive similarity and fuzzy thresholds, and attaches for each span a concise LLM explanation (tertiary label, confidence score, and criterion-referenced rationale). As shown, explanations align closely with the predefined inclusion criteria for \emph{Population}, emphasizing population-level modeling of dementia incidence and mortality and correctly excluding individual-level prediction. This produces an auditable chain from document text to inclusion decision. A legend for label colors and criteria is provided in Table~\ref{tab:legend_for_plots}.

\begin{table}[htbp]
\caption{Lookup table for label colors and inclusion/exclusion criteria used in plots.}
\centering
\begin{adjustbox}{max width=\textwidth}
\begin{tabular}{
    |>{\raggedright\arraybackslash}p{2.5cm}
    |>{\raggedright\arraybackslash}p{2.5cm}
    |>{\raggedright\arraybackslash}p{3.7cm}
    |>{\raggedright\arraybackslash}p{3.7cm}
    |>{\raggedright\arraybackslash}p{3.7cm}|
}
\toprule
\textbf{Criteria} & \textbf{Color} & \textbf{NCD Protocol Criteria} & \textbf{NCD AI Protocol Criteria} & \textbf{Exclusion Criteria} \\
\midrule
Population & Yellow &
Population - Populations at risk of developing NCDs, as defined by the protocol. &
Studies focusing on population-level outcomes (e.g., trends, projections, risk factors at population level). &
Studies focusing on individual-level predictions, or limited subgroups not representative of the population. \\
\midrule
Intervention & Light green &
Intervention, exposure, or scenario (includes both interventions and exposures of interest, as specified in protocol). &
Assessing the impact of hypothetical health risks, interventions, or exposures at a population level. &
Observational studies without simulation, or those not focused on intervention/exposure scenario comparisons. \\
\midrule
Outcome & Light blue &
Outcome - Selected non-communicable diseases (NCDs), as per protocol. &
Comparing the above interventions to current practice or other comparators, with outcomes measured at the population level. &
Studies not involving comparative scenarios or not reporting population-level outcomes. \\
\midrule
Study Design & Light red &
Study approach - Computational simulation models (microsimulation, Markov, system dynamics, agent-based, etc.) &
Outcomes related to the 5 primary NCDs. This includes incidence, prevalence, mortality, or other health outcomes at the population level. &
Studies focusing exclusively on infectious diseases, or study designs outside the scope (e.g., clinical trials, case studies). \\
\midrule
Integration of multiple data sources & Light yellow &
Integration of multiple data sources. Studies incorporating data from multiple sources as defined by the protocol. &
Metrics like all-cause mortality, QALYs, DALYs, life expectancy, or NCD-related economic outcomes. &
Studies not reporting NCD-related metrics or failing to integrate multiple data sources as per the protocol. \\
\bottomrule
\end{tabular}
\end{adjustbox}
\label{tab:legend_for_plots}
\end{table}

\section{Discussion}

The accelerating pace of biomedical publishing makes fully manual systematic reviews increasingly untenable. With thousands of new articles weekly, review teams must triage large evidence bases and, at full text, determine \emph{where} decisive information resides and \emph{how} to interpret heterogeneous, context-dependent language. Our results show that a pipeline expressly designed for this setting, namely contrastive highlighting, fuzzy aggregation, and explainable LLM adjudication, recovers high recall while maintaining traceability and minimizing reviewer effort.

\subsection{Operational Impact and Cost Efficiency}

\begin{figure}[htbp]
    \centering
    \includegraphics[width=\textwidth]{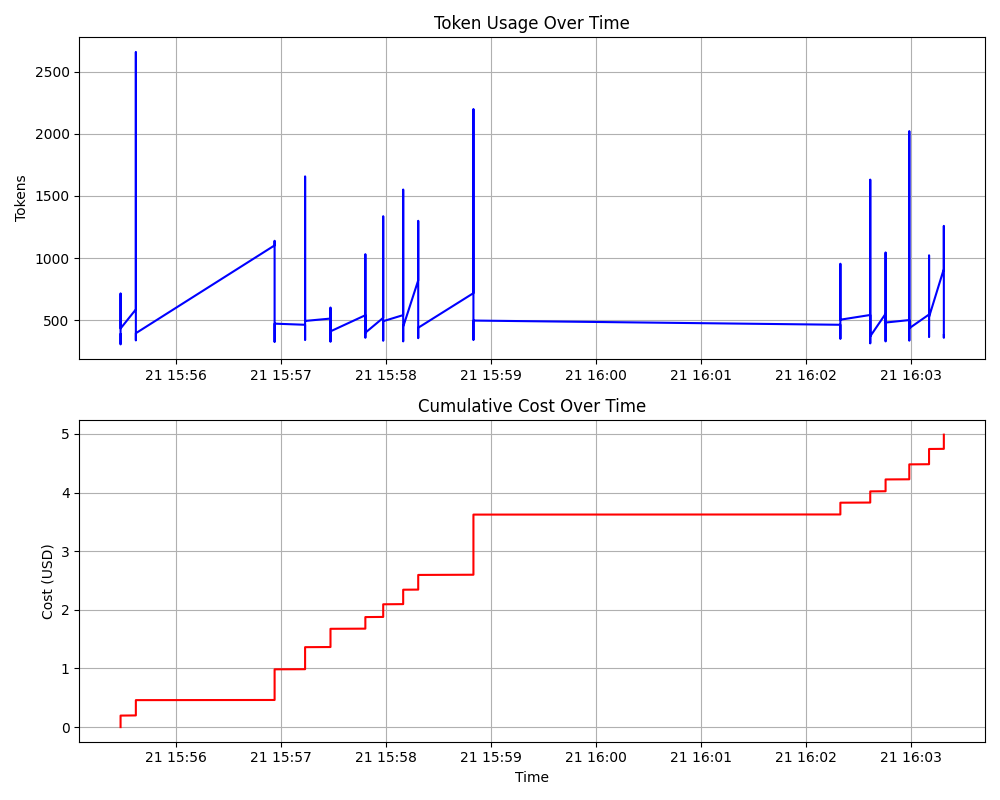}
    \caption{\textbf{Token usage and cumulative cost tracking during automated full-text screening.} The system processes approximately 200 documents for under \$5 USD in total API cost. The integrated cache eliminates redundant API calls on repeated runs, further minimizing cost and accelerating performance.}
    \label{fig:cost_tracker}
\end{figure}

Automating both initial triage and deep full-text screening compresses timelines from months to days, enabling “living” updates at a cadence compatible with policy and guideline work~\cite{Dennstadt2025implement}. In practice, the system functions as a reliable first-pass reviewer that escalates only ambiguous borderline cases to humans, concentrating expert time where it matters most. Cost tracking indicates approximately 200 documents can be processed for under \$5 USD in LLM API fees (Figure~\ref{fig:cost_tracker}). An in-memory cache avoids redundant calls on repeat analyses, yielding negligible marginal cost and near-instant reruns—important for iterative protocol refinement and audit cycles.

\subsection{Protocol Fidelity and Operating Characteristics}

A core requirement for SR automation is strict adherence to protocol definitions. By explicitly contrasting inclusion and exclusion texts, the pipeline anchors decisions in expert-curated criteria. The fuzzy layer converts continuous semantic evidence and explicit vagueness (margin) into graded membership, avoiding the brittleness of crisp rules and the opacity of threshold-only probabilistic scoring. On our all-positive gold set, document-level recalls favored the fuzzy approach over statistical and crisp baselines across criteria, and strict “all-criteria” inclusion rates reflected the same advantage. The design also models \emph{lack of decisive evidence}: if a highlighted span is later judged insufficient by the LLM, its contribution is attenuated rather than forcing hard exclusion, preserving recall on borderline cases while still de-prioritizing weak evidence.

Stability and face validity of explanations are supported by cross-checks: a stronger model (GPT-5) agreed with GPT-4.1-mini on 625/636 commented chunks (98.27\%), and human review agreed on 611/636 (96.07\%). We interpret these as indicators of robustness in the rationale generation and filtering scheme, not as substitutes for gold-standard evaluation.

\subsection{Transparency, Explainability, and Auditability}

Adoption hinges on auditability. Every stage—chunking, embeddings, contrastive scores, fuzzy rule activations, defuzzified $\mu$, LLM labels and rationales—is logged with text offsets, enabling reconstruction of the reasoning chain from document span to decision. The explanations double as training material for new reviewers and support quality assurance and dispute resolution. This traceability directly addresses common barriers to deploying AI in SR workflows and is difficult to realize with conventional black-box classifiers.

\subsection{Scalability and Generalization}

The pipeline is modular and model-agnostic. Criteria texts can be revised without code changes; the fuzzy controller’s universes and rules are interpretable; the LLM layer is batched and swappable. These traits facilitate rapid retargeting to new review topics and maintenance of “living” reviews. Beyond biomedical SRs, the same pattern—contrastive highlighting $\rightarrow$ fuzzy aggregation $\rightarrow$ explainable adjudication—applies wherever diffuse, context-dependent evidence is scattered across long documents (e.g., environmental policy assessments, pharmaco-economic evaluations, social science syntheses).

\subsection{Threats to Validity}

\textbf{Label composition.} The present test set contains only gold positives; as noted in Results, specificity and threshold-free discrimination metrics (ROC/PR curves and AUCs) are undefined here. Mixed-label benchmarks will enable full operating-characteristic reporting, calibration analysis, and paired comparisons.

\textbf{Agreement vs.\ ground truth.} High cross-model and human agreement (98.27\% and 96.07\%) indicate stability and face validity but do not guarantee correctness; shared biases or correlated errors are possible.

\textbf{Domain shift.} Generalization to highly atypical article structures or subdomains with idiosyncratic terminology warrants further evaluation; ablations on highlighting, margin ranges, and rule shapes will help quantify sensitivity.

\textbf{Human factors.} Reviewer behavior may change when assisted by explanations (anchoring, confirmation); prospective human-in-the-loop studies are needed to measure net effects on accuracy and efficiency.

\subsection{Limitations and Future Work}

Future work will (i) evaluate on larger, mixed-label corpora with pre-registered thresholds and report ROC/PR AUCs and calibration; (ii) conduct decision-curve and workload analyses to quantify reviewer effort saved at fixed recall; (iii) perform ablations of highlighting, margin, and fuzzy rules; and (iv) probe disagreement cases to refine prompts, universes, and defuzzification cutoffs. We will also assess portability across domains and LLM back-ends and explore lightweight confidence calibration of LLM judgments. Finally, while specificity could not be assessed here due to the absence of negative annotations, we will consider it in future work as more fully annotated datasets become available.

\subsubsection{Ethical and Privacy Considerations}

In this proof-of-concept, all experimentation used open-access full-text research papers as source material. Patient information is not a concern in this setting, as no identifiable health records are processed. We recognize that potential infringement risks concern both the underlying content and editorial access rights, particularly when such materials are used with GPT-based systems. To mitigate these risks, we limit inputs to licensed or open-access papers when interacting with non-local GPT models.

The system operates as a secure integrated workflow, combining semantic search over a structured database with LLM-based question-resolution and analytical modules, without ever fine-tuning or training its models on copyright-protected full texts. Instead, it employs a retrieval-augmented generation (RAG) approach in which relevant excerpts are retrieved only as a temporary, ephemeral context for a given query. These excerpts are (i) used in real time to answer a specific question, (ii) never stored in the model or reused outside that session, and (iii) never redistributed or published verbatim. Meaningful human creative input is incorporated into any AI-generated outputs.

All content is processed under controlled access, and decisions and rationales are logged for audit. No copyright material is repurposed as training data, and all processing respects licensing terms and editorial rights. Responsible use requires clear provenance, versioning of criteria and models, and reproducible reporting to meet regulatory, ethical, and institutional expectations.

\section{Conclusion}

Full-text screening in systematic reviews is intrinsically \emph{fuzzy}: decisive evidence is scattered across sections, expressed heterogeneously, and rarely admits crisp, binary rules. We reframed the task accordingly and introduced a pipeline that (i) highlights criterion-relevant spans via contrastive semantic similarity, (ii) aggregates evidence with a transparent Mamdani fuzzy controller, and (iii) provides auditable, criterion-referenced rationales through an LLM judge while conservatively handling the \emph{lack of evidence} case. This design couples high recall with end-to-end traceability and practical efficiency.

On an all-positive gold set ($N{=}16$ documents; $M{=}3{,}208$ chunks), the fuzzy approach outperformed statistical and crisp baselines in document-level recall across Population, Intervention, Outcome, and Study Approach, and doubled the strict “all-criteria” inclusion rate relative to a crisp baseline. Explanations were stable under a stronger model (GPT-5 agreement: 98.27\%) and aligned with human inspection (96.07\%), while a small reviewer study found high agreement (91\%, $\kappa{=}0.82$) and a reduction of per-article screening time from $\sim$20 minutes to under a minute. The system is inexpensive to operate at scale (about \$5 USD per $\sim$200 documents), and all decisions are logged from span to summary, enabling audit, training, and dispute resolution.

Fuzzy logic proved to be a natural formalism for full-text inclusion/exclusion: it preserves graded signals, exposes its rule activations, and integrates cleanly with modern NLP and LLMs. Beyond biomedical SRs, the same blueprint—contrastive highlighting $\rightarrow$ fuzzy aggregation $\rightarrow$ explainable adjudication—applies wherever long, heterogeneous documents embed diffuse, context-dependent evidence.

This work has limits. The present evaluation emphasizes recall on an all-positive set; full operating characteristics (specificity, ROC/PR curves and AUCs, calibration) require mixed-label benchmarks and larger, prospective studies. Agreement analyses indicate stability and face validity but do not replace gold-standard adjudication, and robustness under domain shift merits further testing. Future work will therefore expand to mixed-label corpora with pre-registered thresholds, conduct ablations on highlighting/margin/rule universes, assess reviewer workload via decision-curve analysis, and perform bias/robustness audits. 

Taken together, our results suggest that fuzzy, explainable screening can transform full-text review from a brittle, labor-intensive bottleneck into a transparent, scalable process suitable for living reviews and institutional deployment. Our open-source implementation provides a practical starting point for reproducible adoption and community refinement.


\end{document}